\documentclass{article}

\usepackage{arxiv}

\usepackage[utf8]{inputenc} 
\usepackage[T1]{fontenc}    
\usepackage{hyperref}       
\usepackage{url}            
\usepackage{booktabs}       
\usepackage{amsfonts}       
\usepackage{nicefrac}       
\usepackage{microtype}      
\usepackage{lipsum}		
\usepackage{graphicx}
\usepackage{natbib}
\usepackage{doi}
\usepackage{multirow}

\title{Wheat Head Counting by Estimating a Density Map with Convolutional Neural Networks}

\date{January 10,2021}	
\author{{Hongyu Guo}\\
	Division of Biomedical Engineering\\
	University of Saskatchewan\\
	Saskatoon, SK, Canada, S7N 5A2\\
	\texttt{hpg812@usask.ca} }

\renewcommand{\shorttitle}{\textit{Wheat Head Counting} }

\hypersetup{
pdftitle={Wheat Head Counting by Estimating a Density Map with Convolutional Neural Networks},
pdfsubject={q-bio.NC, q-bio.QM},
pdfauthor={Hongyu Guo},
pdfkeywords={Wheat head counting, Object counting, Density map, Convolutional neural network, deep learning}
}

\begin{document}
\maketitle

\begin{abstract}
Wheat is one of the most significant crop species with an annual worldwide grain production of 700
million tonnes. Assessing the production of wheat spikes can help us measure the grain production. Thus, detecting and characterizing spikes from images of wheat fields is an essential component in a wheat breeding process. In this study, we propose three wheat head counting networks (WHCNet\_1, WHCNet\_2 and WHCNet\_3) to accurately estimate the wheat head count from an individual image and construct high quality density map, which illustrates the distribution of wheat heads in the image. The WHCNets are composed
of two major components: a convolutional neural network (CNN) as the front-end for wheat head image feature extraction and a CNN with skip connections for the back-end to generate high-quality density maps. The dataset used in this study is the Global Wheat Head Detection (GWHD) dataset, which is a large, diverse, and well-labelled dataset of wheat images and built by a joint international collaborative effort. We compare our methods with CSRNet, a deep learning
method which developed for highly congested scenes understanding and 
performing accurate count estimation as well as presenting high
quality density maps. By taking the advantage of the skip connections between CNN layers, WHCNets integrate features  from low CNN layers to high CNN layers, 
thus, the output density maps have both high spatial resolution and detailed representations of the input images. 
The experiments showed that our methods outperformed CSRNet in terms of  the evaluation metrics, mean  absolute  error  (MAE)  and  the  root  mean squared  error  (RMSE) with smaller model sizes. The code has been deposited on GitHub (\url{https://github.com/hyguozz}).
\end{abstract}


\section{Introduction}
Wheat is an important primary food for a large proportion of the world's population, so methods to estimate and enhance its yield have received significant research attention\cite{bogn1}. 
Genomic selection and high-throughput phenotyping techniques are essential in selecting important wheat traits, such as, yield potential, disease resistance, or adaptation to abiotic stress. Developing efficient and robust models for traits extraction of raw data is challenging\cite{wheat1}. 
Wheat head density, the number of wheat heads per unit ground area, is a significant yield trait. However, wheat head counts and density is mainly dependent manually evaluation, which is labour intensive and inaccuracy resulting in around 10\% measurement errors\cite{madec1}.
Therefore, automated wheat head detection and counting methods based on machine learning technology
can help to estimate wheat yield and discover the potential traits of wheat phenotyping\cite{zhang}. 
The computer vision based object counting task is the estimation of the number of objects presented in images or videos. Since the potential wide range of real-world applications such as public safety, traffic
control, agriculture monitoring, and cell counting, object counting has been extensively explored by many researchers\cite{Jiang_2020_CVPR}\cite{Lempitsky}. Object detection methods can  localize and identify
wheat heads in images, so the head density
of wheat populations can be estimated. Wheat head counting can also discover the additional wheat traits, including
the spatial distribution between rows, 
the presence of
awns, size, inclination, colour, 
grain filling stage, and health. Thus, wheat head counting could help farmers to manage their crops scientifically\cite{wheat1}.

The counting task of the number of objects in an image can be classified into two categories, counting by detection and counting by regression\cite{Lempitsky}\cite{Segui_2015}.
Counting by detection uses a object detector to localize individual objects in the image. Given the bounding boxes of all instances or a single dot on each object instance
in each image\cite{Lempitsky}, counting can be easily performed.
However,
object detection is very far from being solved. The extreme overlap of objects, the size of the instances, scene perspective, etc. can affect the performance of the supervised object counting systems, thus, many researches have turned their attentions to the density map based object counting models\cite{arteta2014interactive}\cite{zhang2015cross}\cite{Lempitsky}
. Density map based object counting methods tackle the counting problem through learning a regression function that projects the image appearance into an object density map, then obtain the object count by integration. Moreover, density map preserves more information and gives the spatial distribution of the objects in a given image\cite{Jiang_2020_CVPR}. 

However, for the supervised counting methods, the quality of the annotations of images is crucial in deciding the accuracy of the counting task, besides, object detection requires time-consuming inference. To avoid the these disadvantages, unsupervised object counting methods were proposed, which tackle counting problems based
on grouping self-similarities or motion similarities, thus avoid the complicated object detection processing\cite{Rabaud}\cite{4408926}. 

Image-based plant phenotyping has received increasing interest in recent years. 
Organ counting is a common task in image-based plant phenotyping, such as leaf, head, pod, fruit counting etc. Tewodros Ayalew et al.\cite{ayalew2020unsupervised} proposed a
domain-adversarial learning approach for domain adaptation of density
map estimation for the purposes of object counting and evaluated on wheat spikelets counting and leaves counting. 
Jordan et al.\cite{ubbens2020autocount} used a fully unsupervised method to implement the plant organ counting task, which is a convolutional network-based unsupervised segmentation method followed by two post-hoc optimization steps. Bipul Neupane et al.\cite{bana} proposed a deep learning based method to detect and count banana plants on a farm exclusive of other plants, using high resolution RGB aerial images collected from Unmanned Aerial Vehicle (UAV). Since the available datasets of the plant phenotyping  are often small and the costs associated with generating new data are high. Ubbens et al.\cite{Ubbens} proposed a method for augmenting plant phenotyping datasets using rendered images of synthetic plants and estimated their method on CNN based leaf counting task. 

In this study, we take a supervised learning approach to solve the wheat head counting problem, so require a set of training wheat head images with annotations. Global Wheat Head Detection (GWHD) dataset\cite{wheat1} is a large, diverse, and well-labelled dataset of wheat images and built by a joint international collaborative effort. The GWHD dataset is publicly available at \href{URL}{http://www.global-wheat.com/}. Based on the GWHD dataset, we propose three deep learning based wheat head counting networks (WHCNet\_1, WHCNet\_2 and WHCNet\_3) to detect and count wheat spikes presented in wheat field images. The WHCNets are composed
of two major components: a CNN based front end for wheat head image feature extraction and a CNN with skip connections for the back end to generate high-quality density maps, consequently,  accomplish the wheat head counting task.
\section{Methods}
Density map based wheat head counting refers to the input is a wheat head image and the output is the density map of the wheat heads, which shows how many wheat heads per unit area and the spatial distribution of wheat heads in that image, so it is very useful in many applications, such as, estimating the grain yield potential. Consequently, the number of wheat heads in an image can be obtained by the integration of its density map.
In this section, firstly, we will introduce the dataset and data preprocessing, then, we will discuss how to generate the ground truth density maps from wheat head images. For comparison purpose, we introduce a baseline network, CSRNet, which has achieved the state-of-the art performance on dense crowd counting tasks and vehicle counting tasks. Consequently, we present three wheat head counting networks, WHCNet\_1, WHCNet\_2 and WHCNet\_3, which can learn density maps from input wheat head images via fully CNNs. The loss function and evaluation metrics will be described as well.
\subsection{Dataset and data preprocessing}
Global wheat head detection (GWHD) dataset\cite{wheat1} is collected from several countries around the world at different growth stages with a wide range of genotypes aiming at developing and benchmarking methods for wheat head detection. In terms of phenotyping datasets for object detection, 
GWHD dataset is currently the largest open labelled dataset freely available for object detection for field plant phenotyping. 

\begin{figure}
\centering
    \includegraphics[width=0.90\columnwidth]{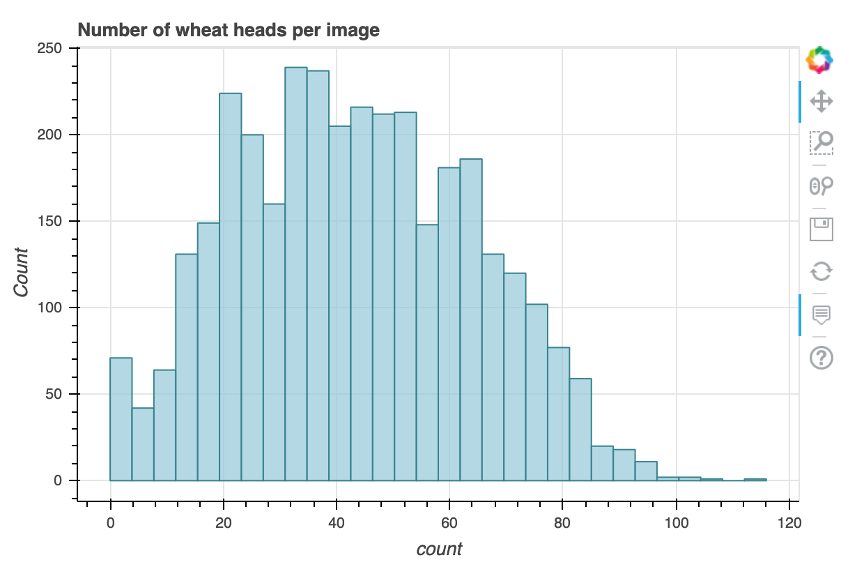}
    \caption{The count number of bounding boxes per image. }
    \label{fig:count}
\end{figure}

In this study, we downloaded the GWHD dataset from \href{URL}{https://www.kaggle.com/c/global-wheat-detection}, which contains 3422 high-resolution RGB images for training and 10 high-resolution RGB images for testing, with 147793 wheat head with bounding boxes annotated which average
40 heads per image. 
Figure \ref{fig:count} shows the distribution of the count number of bounding boxes per image. As can be seen from the figure, most of the images have 20-60 wheat heads, and few images, specifically 4 images, contain more than 100 heads with a maximum of 116 heads. Moreover, there are 49 images containing no heads in the dataset. 

As deep learning framework requires a large amount of training data.
We crop four patches from four angles of each image with 1/4 size of the original image. Then, we vertical flip the patches to further double the wheat head image dataset, thus increase the size of our training set by a factor of 8. 
We have mentioned that in the GWHD dataset (kaggle version), there are 
3422 images in the training folder and 10 images in testing folder. We only augmented the images of the training folder and split these patches into training set, validation set and testing set. For the 10 images in the testing folder, as there are no annotation information given, we leave them out and use them to verify the performance of our models. 
Since the GWHD dataset has sub-datasets from different regions in the world, we shuffle these augmented image patches to ensure the images from different regions distributed in our training set, validation set and testing set, evenly. As a result, we 
 selecte 12,000 patches for training, 1,600 patches for validation set and 1,600 patches as the testing set. 
\subsection{Ground truth density map generation}
The wheat head counting solution requires a set of annotated wheat head images, where all the wheat heads are marked by dots. The ground truth density map $D_{I}$
, for a wheat head image $I$, is defined as a sum of Gaussian functions centered on each dot annotation,
\begin{equation}
    D_{I}(p)=\sum_{\mu \in A_{I}}N(p;\mu ,\sigma )
\end{equation}
where $A_{I}$ is the set of 2D points annotated for the image $I$, and $N(p;\mu ,\sigma^{2} )$ represents the evaluation of a normalized 2D Gaussian function, with mean $\mu$ and isotropic co-variance matrix $\sigma^{2}$, evaluated at pixel position defined by $p$. With the density map $D_{I}$, the total wheat head count $N_{I}$ can be directly obtained by integrating the density map values in $D_{I}$ over the entire wheat head image, as follows,
\begin{equation}
    N_{I}=\sum_{p\in I}D_{I}(p).
\end{equation}
Since all the Gaussian are summed, so the total wheat head count is preserved even when there is overlap between wheat heads. The purpose of our wheat head counting model is to learn a mapping from the input wheat head image to a wheat head density map. 

However, in this definition of density function, each object is looked as independent samples in the image, thus, the perspective distortion, and
the pixels associated with different samples correspond
to areas of different sizes in the scene are all neglected. The  geometry-adaptive kernels\cite{MCCNN} takes the distortion caused by the homography
between the ground plane and the image plane into account by assuming around each object, the objects are 
evenly distributed, then the average distance between
this object and its nearest $k$ neighbors (in the image) gives a
reasonable estimate of the geometric distortion (caused by
the perspective effect). Therefore, for the
density maps of those dense scenes, the spread parameter $\sigma$ for each object can be determined 
based on its average distance to its neighbors, fromalized by 

\begin{equation}
     \sigma _{i}=\beta \bar{d_{i}}
    \label{eq:geo}
\end{equation}
where $\bar{d_{i}}$ represents the average distance of $k$ nearest neighbors of the $i$th object. Thus, the Gaussian kernel with variance $\sigma _{i}$ is proportional to $\bar{d_{i}}$, and $\beta$ is a regulating parameter. 

\begin{figure}
    \centering
    \includegraphics[width=0.9\columnwidth]{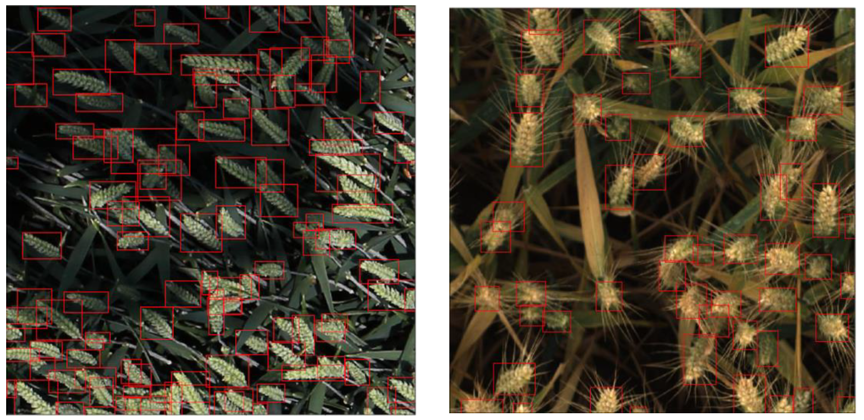}
    \includegraphics[width=0.9\columnwidth]{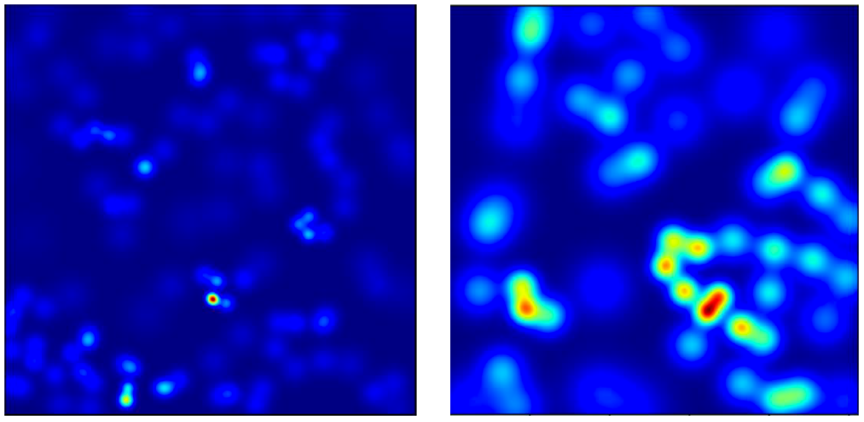}
    \caption{Wheat head images with dense wheat head density, medium dense wheat head density. The left side shows the bounding box labeled wheat head images. The right side shows the corresponding density maps generated by geometry-adaptive kernel with $\beta=0.3, k=3$.}
    \label{fig:density}
\end{figure}

In this study, we adopt the geometry-adaptive kernels to generate the ground truth of wheat head images because most of the wheat heads are densely distributed in our images, similar as the dense crowd scene in the study of \cite{Li_2018}. As the GWHD dataset has provided the bounding box annotations, firstly, the dot-annotation can be obtained through calculating the centroid of each bounding box, then, the ground truth density maps for all wheat head images are generated. 
Figure \ref{fig:density} shows the bounding box labeled wheat head images, and their corresponding density maps generated using  the centroids of bounding boxes. The $\beta$ is set as 0.3, and $k$ is set as 3, followed the configuration in paper \cite{Li_2018}, as the wheat heads image is a dense image, similar with the context of crowds counting problem\cite{Li_2018}. 
\subsection{Baseline network}
Li et al.\cite{Li_2018} proposed CSRNet for congested scene recognition, which can perform crowd count estimation and generate high quality density maps. CSRNet obtained the state-of-the-art performance  in four crowd counting datasets (ShanghaiTech dataset, the UCF CC 50 dataset,
the WorldEXPO’10 dataset, and the UCSD dataset)\cite{MCCNN} compared with other  previous state-of-the-
art methods and achieved the best accuracy on the vehicle counting task as well. 
\begin{figure}
    \centering
    \includegraphics[width=0.9\columnwidth]{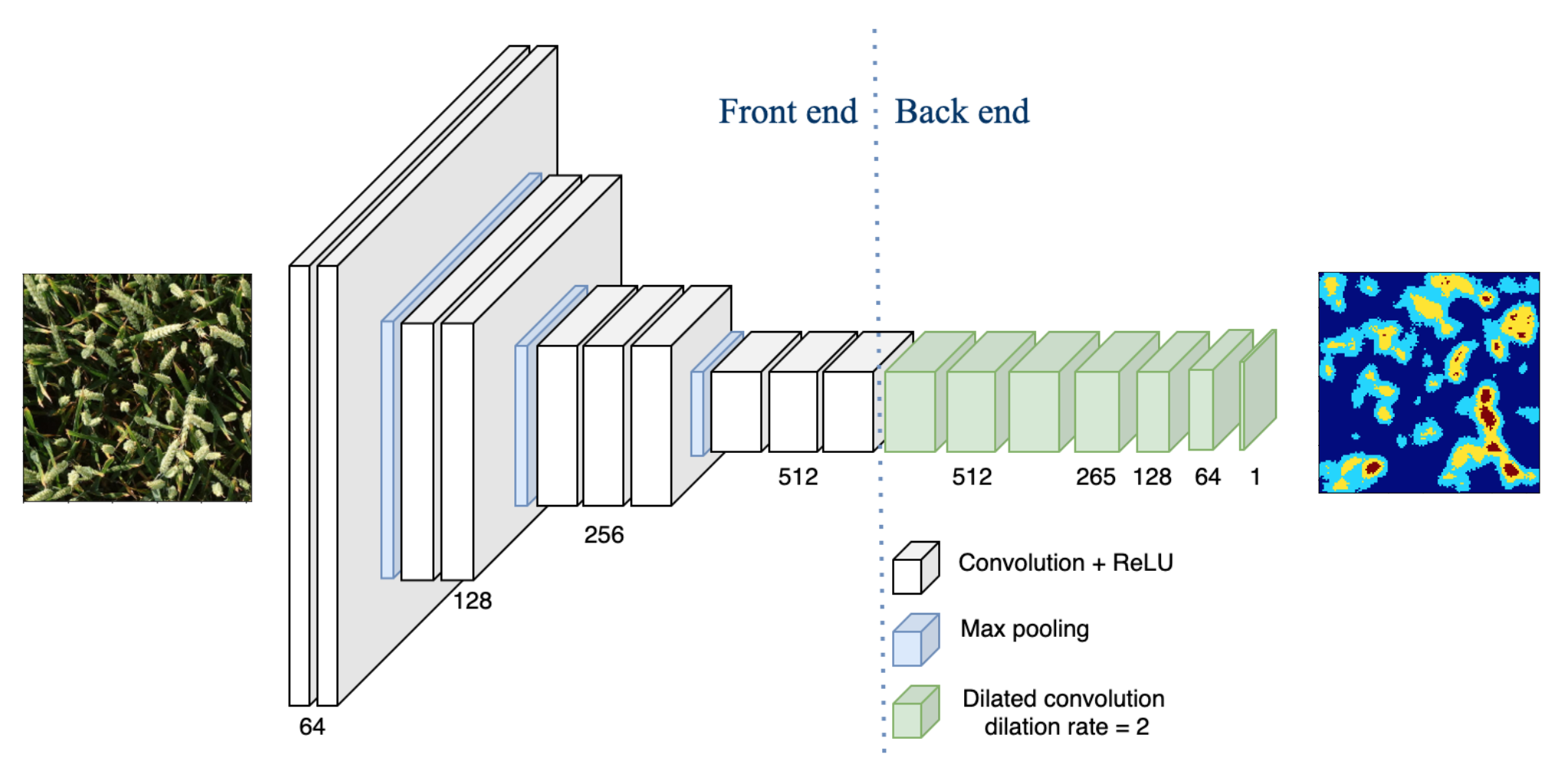}
    \caption{The architecture of the baseline model CSRNet. 
    All convolutional layers use
padding to maintain the previous size. The convolutional layers'
parameters are denoted as '(number of layers) -conv-(kernel size)-(number of filters)-
(dilation rate, if applicable)', max-pooling layers are conducted over a $2 \times 2$
pixel window with stride 2. Thus, the configuration of CSRNet is as follows: Front end ( 2-conv3-64, 1-max-pooling, 2-conv3-128, 1-max-pooling, 3-conv3-256, 1-max-pooling, 3-conv3-512), Back end ( 3-conv3-512-2, 1-conv3-256-2, 1-conv3-128-2, 1-conv3-64-2), Output layer ( 1-conv1-1) }
    \label{fig:abaseline}
\end{figure}
Since in our wheat head images, wheat heads often overlap and occlude each other with high planting densities, the wheat head counting problem is a dense counting problem as well. Therefore, we take CSRNet as the baseline of our study.

Figure \ref{fig:abaseline} illustrates the architecture of CSRNet, as shown in this figure, CSRNet is composed of two parts: a CNN as the front-end and a dilated CNN for the back-end. It is a fully
convolutional network. Specifically, the front-end of CSRNet is composed of the top 10 convolutinal layers and 3 maxpooling layers of VGG-16\cite{simonyan2014deep} while the back-end is a CNN with 6 consecutive convolutional layers. The architecture is concise, however, the high level features extracted from the deeper layers are more abstract and some location information from the lower layers may be lost. 
Even though those abstract high level features can help to improve the performances in some classification tasks where only one class prediction is mapped from the input image, but when the features are used for the generation of density map, the high level features are not enough to construct the location information of the input image. Therefore, the deeper CNN layers in CSRNet can not preserve enough spatial information to generate accurate density maps from the input images. 

\subsection{WHCNet architecture}
\begin{figure*}

    \centering
    \includegraphics[width=0.9\columnwidth]{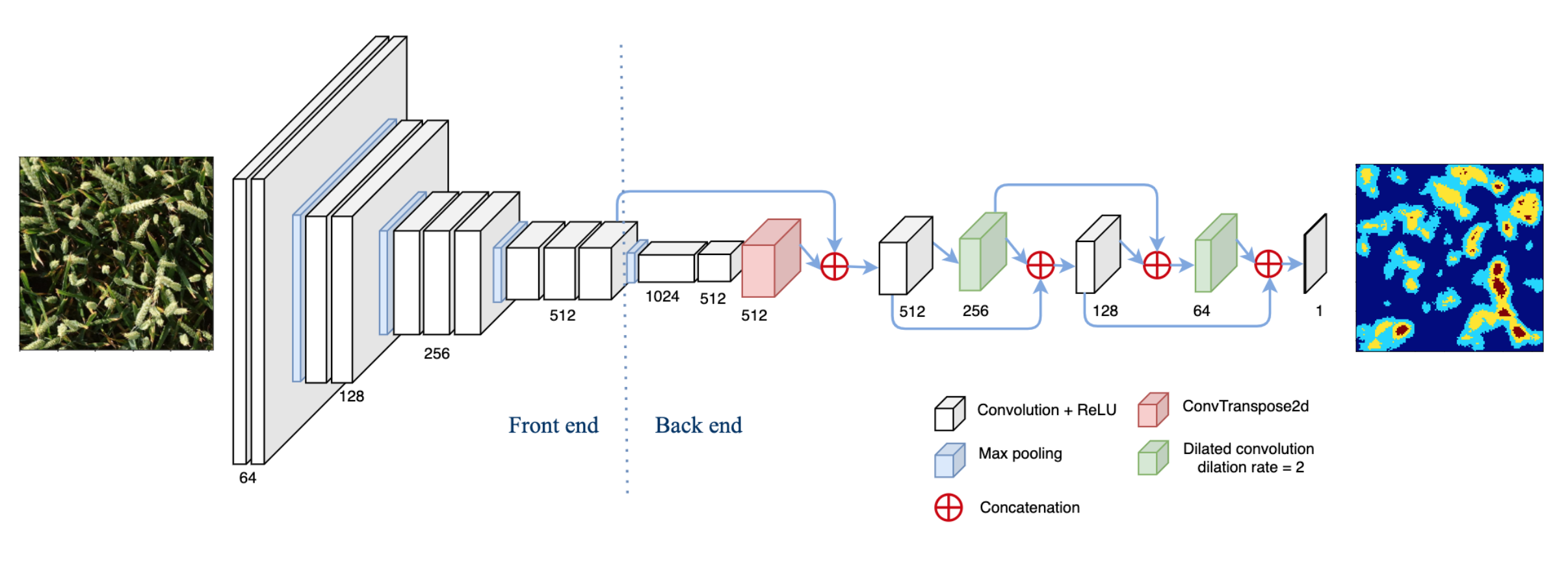}
    \caption{The overall architecture of WHCNet\_1.  
    All convolutional layers use
padding to maintain the previous size. The convolutional layers'
parameters are denoted as '(number of layers) -conv-(kernel size)-(number of filters)-
(dilation rate, if applicable)', max-pooling layers are conducted over a $2 \times 2$
pixel window with stride 2. Thus, the configuration of CSRNet is as follows: Front end ( 2-conv3-64, 1-max-pooling, 2-conv3-128, 1-max-pooling, 3-conv3-256, 1-max-pooling, 3-conv3-512), Back end ( 1-max-pooling, 1-conv3-1024, 1-conv3-512, 1-convtranspose3, 1-conv3-512, 1-conv3-256-2, 1-conv3-128, 1-conv3-64-2), Output layer ( 1-conv1-1) }
    \label{fig:archi}
    
\end{figure*}

\begin{figure*}
    \centering
    \includegraphics[width=0.9\columnwidth]{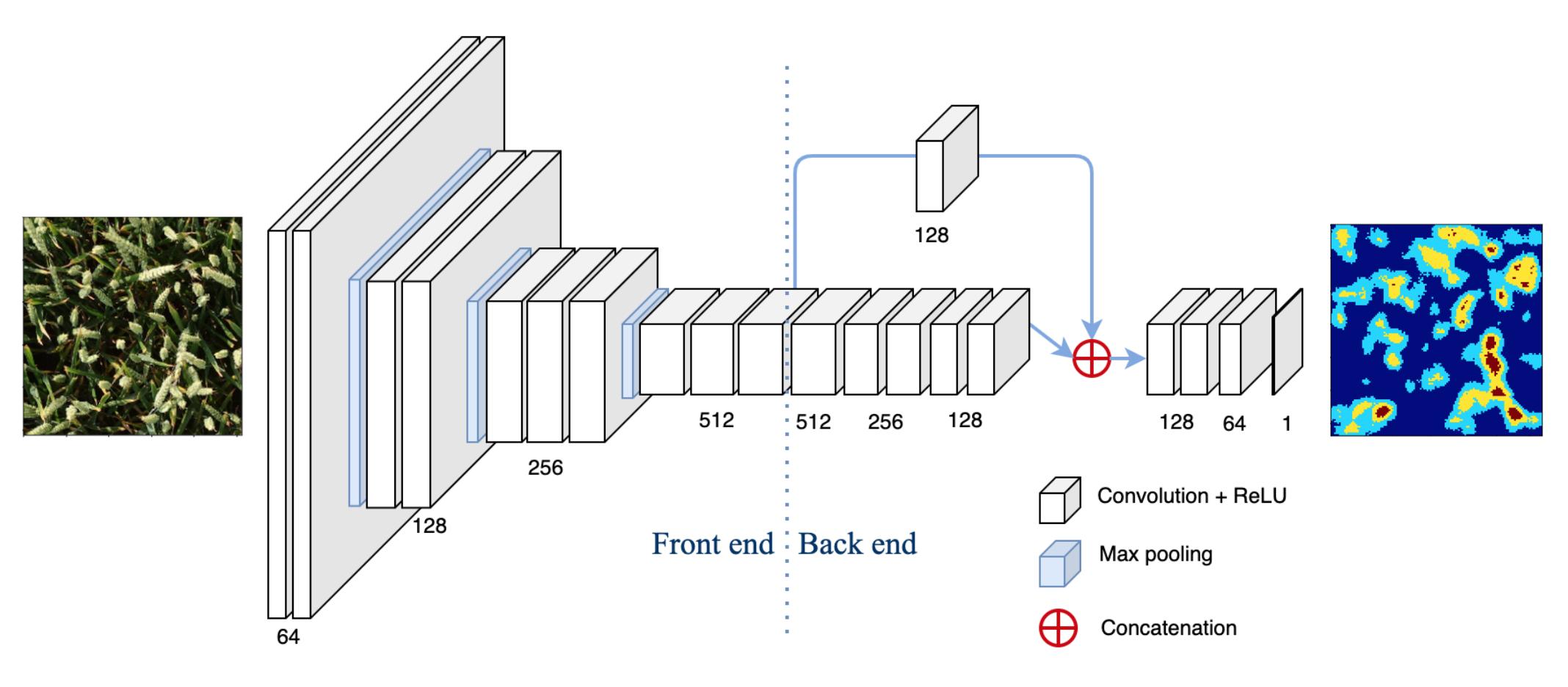}
    \caption{The overall architecture of WHCNet\_2.  
    All convolutional layers use
padding to maintain the previous size. The convolutional layers'
parameters are denoted as '(number of layers) -conv-(kernel size)-(number of filters)-
(dilation rate, if applicable)', max-pooling layers are conducted over a $2 \times 2$
pixel window with stride 2. Thus, the configuration of CSRNet is as follows: Front end ( 2-conv3-64, 1-max-pooling, 2-conv3-128, 1-max-pooling, 3-conv3-256, 1-max-pooling, 3-conv3-512), Back end ( skip connection part [ 1-conv3-512, 2-conv3-256, 2-conv3-128, parallel with 1-conv3-128],  consecutive part[  2-conv3-128, 1-conv3-64]), Output layer ( 1-conv1-1) }
    \label{fig:arch2}
\end{figure*}

As discussed in the last section, the back end of CSRNet may cause the loss of the location information of the input image as the high level features are more abstract than the low level features, thus, the quality of the output density map can be degraded. To address this issue, we introduce forward skip connections into the back end of our models aiming to infuse the location information of wheat heads from low layers to high layers to avoid information degradation during the training process and integrate low layer features with high level features for the inference of density maps. We propose three architectures, WHCNet\_1 (see Figure \ref{fig:archi}),  WHCNet\_2 (see Figure \ref{fig:arch2})  and WHCNet\_3\ref{fig:arch3}, respectively. There are two major components in our models, the front end CNN and the back end CNN. We use the same front end with CSRNet, which includes the top 10 convolutional layers and 3 max-pooling layers of VGG-16 model, to tackle the low level features extraction. Besides, the pre-trained VGG-16 model has been previously trained on a large dataset and contains the weights and biases that represent the features of the dataset it was trained on. Thus, using the  pre-trained VGG-16 model, we can not only save the training time but also obtain more accurate weights instead of using random initialized weights. 


As for the back end part, we propose three architectures, which will be described in the following section. The last layer in WHCNet\_1, WHCNet\_2 and WHCNet\_3 is the  convolutional layer with the kernel size $1\times 1$, which is responsible for outputting the density map of the input image. In addition, the kernel size for all convolution including the transpose convolution is set to $3 \times 3$, which has shown
excellent image recognition performance\cite{simonyan2014deep}. 
Moreover, the rectified linear activation function (or ReLU for short) layers are added after each convolutional layer as it has been demonstrated that it can help the model to be easier to be trained and  achieve better performance. Besides, it is worth mentioning that the size of input image can
be arbitrary since our network is essentially a pixel-wise prediction.
\subsubsection{WHCNet\_1}
Inspired by U-Net\cite{ronneberger2015unet}, an architecture composed of a contracting path to capture context and a symmetric expanding path that enables precise localization. U-Net uses the upsampling part to propagate context information to deep layers. In this study, to further extract features from the input image, we build a downsampling and upsampling part by adding one max-pooling layer with stride 2 at the beginning of the back end part, after two consecutive convolutional layers, an upsampling layer is applied to restore the spatial resolution, subsequently, the output of the deeper layer is concatenated with the low level features, i.e. the output of the front end part, thus we can achieve performance improvement while the network going deeper. Specifically, we use the transpose convolution layer as the upsampling layer as it is a learnable upsampling layer. 

Consequently, consecutive convolutional layers with braided skip connections are used to construct the density map. These skip connections from earlier layers in the network provide the necessary detail to reconstruct accurate shapes for the density maps. In addition, two dilated convolutional layers with dilation rate 2 are included in the back end. 
Dilated convolution (also called Atrous convolution) can arbitrarily enlarge
the field-of-view of filters at deep CNN layer\cite{Yu2016MultiScaleCA}. Dilated convolution
with rate $r$ introduces $r-1$ zeros between consecutive filter
values, thus, enlarging the kernel size of a $k\times k$ filter
to $k+(k-1)(r-1)$ without increasing the number
of parameters. This character enlarges the receptive field without increasing the number of parameters or the amount of computation.
Dilated convolutional layers have been demonstrated having significant improvement of accuracy in segmentation tasks\cite{NIPS2018_8087}. The dilated convolution layers are used in WHCNet\_1 to extract multi-scale representation from wheat head images. 

WHCNet\_1 utilizes the downsampling and upsampling part to extract high level features, a braided skip connection part to provide location information of the input image from lower layers to deeper layers, and the dilated convolutional layers to extract multi-scale features. However, the architecture of WHCNet\_1 is a little bit complicated compared with the baseline CSRNet. To overcome this disadvantage, we propose WHCNet\_2, a simpler architecture compare to  WHCNet\_1 in next section. 
\subsubsection{WHCNet\_2}
In WHCNet\_2 (See Figure \ref{fig:arch2}), we use one skip connection from the output of the front end of the model to the output of 5 stacked convolutional layers  to combine the location information from the low layers with the high level features to ensure the output density map preserve both the higher level representation and accurate location information. Different from other skip connections\cite{7780459}\cite{ronneberger2015unet}, we add one convolutional layer in the connection path aiming to keep the output shape of the front end part consistent with the output shape of its prior consecutive CNN layers, so that the concatenation layer can accept balanced information with the equal portion of the low and high level features.  
\subsubsection{WHCNet\_3}
\begin{figure}
    \centering
\includegraphics[width=0.9\columnwidth]{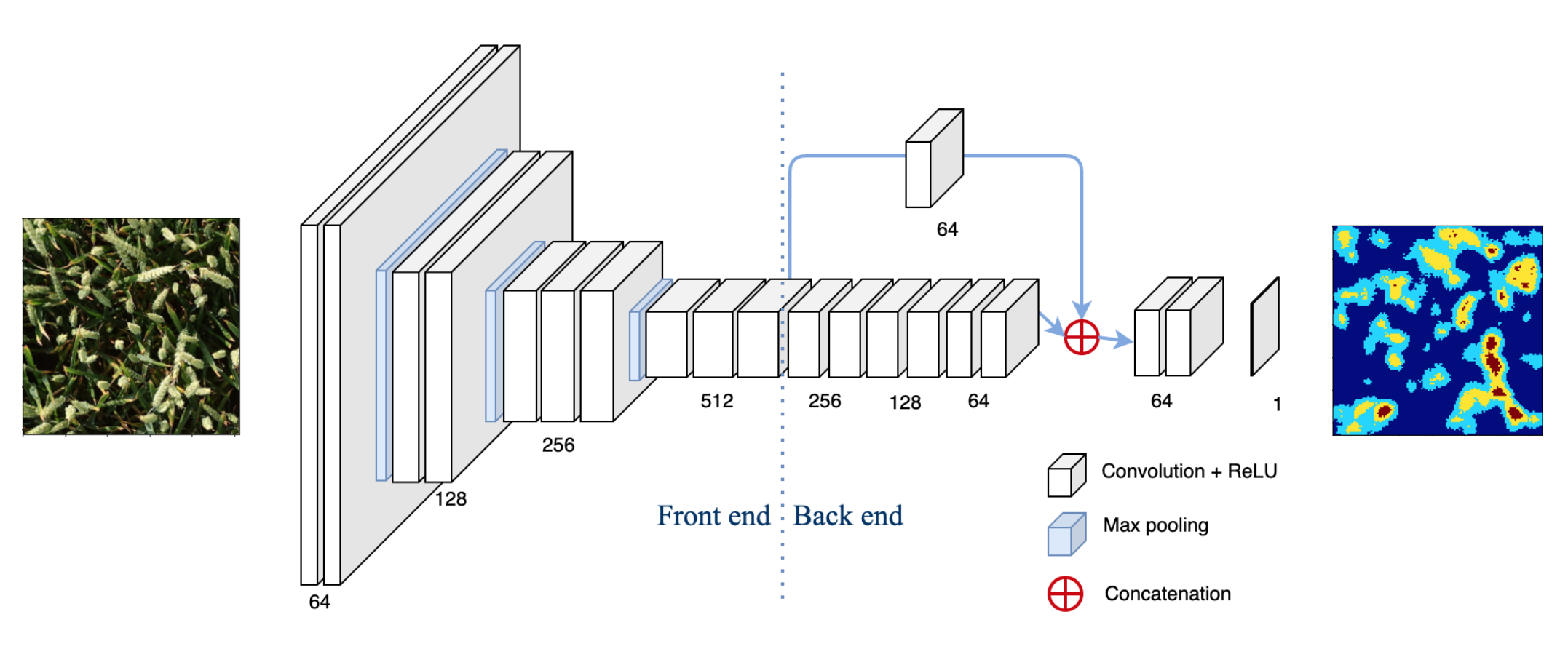}
    \caption{The overall architecture of WHCNet\_3.  
    All convolutional layers use
padding to maintain the previous size. The convolutional layers'
parameters are denoted as '(number of layers) -conv-(kernel size)-(number of filters)-
(dilation rate, if applicable)', max-pooling layers are conducted over a $2 \times 2$
pixel window with stride 2. Thus, the configuration of CSRNet is as follows: Front end ( 2-conv3-64, 1-max-pooling, 2-conv3-128, 1-max-pooling, 3-conv3-256, 1-max-pooling, 3-conv3-512), Back end ( skip connection part [ 2-conv3-256, 2-conv3-128, 2-conv3-64, parallel with 1-conv3-64],  consecutive part[ 2-conv3-64]), Output layer ( 1-conv1-1) }
    \label{fig:arch3}
\end{figure}

To further reduce the computation cost WHCNet\_2, we design WHCNet\_3, with the similar architecture of WHCNet\_2, but fewer convolutional layers and smaller filter sizes for convolutional layers. The architecure of WHCNet\_3 is shown in Figure \ref{fig:arch3}. 
\subsection{Loss function}
Euclidean distance is used to measure the difference between the estimated density map and ground truth\cite{Li_2018}\cite{MCCNN}\cite{shi2019counting}. The loss function is defined as follow:

\begin{equation}
    L(\Theta )=\frac{1}{2N}\sum_{i=1}^{N}\left \| \Psi (Img_{i};\Theta)- GT_{i} \right \|_{2}^{2}
\end{equation}
where $\Theta$ is a set of trainable parameters in our deep CNN network $\Psi$, $N$ is the number of the training images in the batch, $Img_{i}$ is the input wheat head image and $GT_{i}$ is the corresponding ground truth density map. Therefore, $L$ is the loss between the estimated density map and the ground truth density map.
\subsection{Evaluation metrics}
The mean absolute error (MAE) and the root mean squared error (RMSE) between the predicted and ground truth maps are used for the evaluation of the counting performance in this study, which are defined as follows:
\begin{equation}
    MAE=\frac{1}{N}\sum_{i=1}^{N}\left | Cest_{i}- Cgt_{i}\right |
\end{equation}
\begin{equation}
    RMSE=\sqrt{\frac{1}{N}\sum_{i=1}^{N}\left | Cest_{i}-Cgt_{i}\right |^{2}}
\end{equation}
where $N$ is the number of test images, $Cest_{i}$ stands for the estimated counting number of the $i$th image $Img_{i}$, and $Cgt_{i}$ represents the corresponding ground truth of counting. 
\section{Results} 
We present experiments trained on the training set, which is composed of 12000 patches of wheat head images, and validated on the validation set, which has 1600 patches of wheat head images,  using models, CSRNet, WHCNet\_1, WHCNet\_2 and WHCNet\_3, separately. We test these four models on the 1600 patches (size of $512 \times 512$) testing set, and we also test the models on the full sized wheat head images (size of $1024 \times 1024$) by randomly selecting 350 wheat head images from GWHD.   

Implementation of the proposed networks and their experiments are
based on the Pytorch framework. The initialisation of the network weights is important, since bad initialisation consumes time to  the learning process due to the instability of gradient in deep nets. We use the pretrained VGG-16 model as the front-end in this study. For the other layers, we use Gaussian initialization with 0.01 standard deviation to initialize our models. Stochastic gradient
descent (SGD) is applied with fixed learning rate at 1e-6 
during training. 

The experiments results are compared in Table \ref{tab:res}.

\begin{table}
\centering
\caption{Comparison of the performances of WHCNets and the baseline model CSRNet.}
\label{tab:res}
\begin{tabular}{cccccc}
\hline
\\Model & MAE          & RMSE         & MAE            & RMSE           & Size
\\ \cline{2-5}
                       & \multicolumn{2}{c}{Patches} & \multicolumn{2}{c}{Whole image} &                       \\ \hline
WHCNet 1               & \textbf{1.895}        & \textbf{2.431}        & \textbf{6.251}          & \textbf{7.972 }         & 202 M                 \\
WHCNet 2               & 1.955        & 2.523        & 6.329          & 8.089          & 102 M                 \\
WHCNet 3               & 2.01         & 2.591        & 6.627          & 8.43           & \textbf{79 M }                 \\
CSRNet                 & 2.352        & 3.006        & 7.923          & 9.911          & 124 M                 \\ \hline
\end{tabular}
\end{table}
As shown in Table \ref{tab:res}, the model WHCNet\_1 has achieved the best MAE and RMSE both in testing patches and the whole images, however, its model size is the biggest one in the four models, because WHCNet\_1 includes a down sampling and up sampling part, and four skip connections, the architecture is complicated compared with other models. To reduce the complexity and the cost of computation, we removed the down and up sampling part and optimized the skip connection part to keep only one skip connection in WHCNet\_2 and WHCNet\_3, therefore their model sizes have decreased, meanwhile, the performances dropped slightly in terms of the evaluation metrics of MAE and RMSE. Though WHCNet\_2 and WHCNet\_3 have the smaller model sizes than CSRNet, our proposed models outperformed the baseline model CSRNet at the evaluaiton metrics MAE and RMSE. This has testified that the skip connection scheme can improve the performance compared with the consecutive layers scheme in our wheat head counting task. 

\begin{figure}[hbt!]
    \centering
    \includegraphics[width=0.9 \columnwidth]{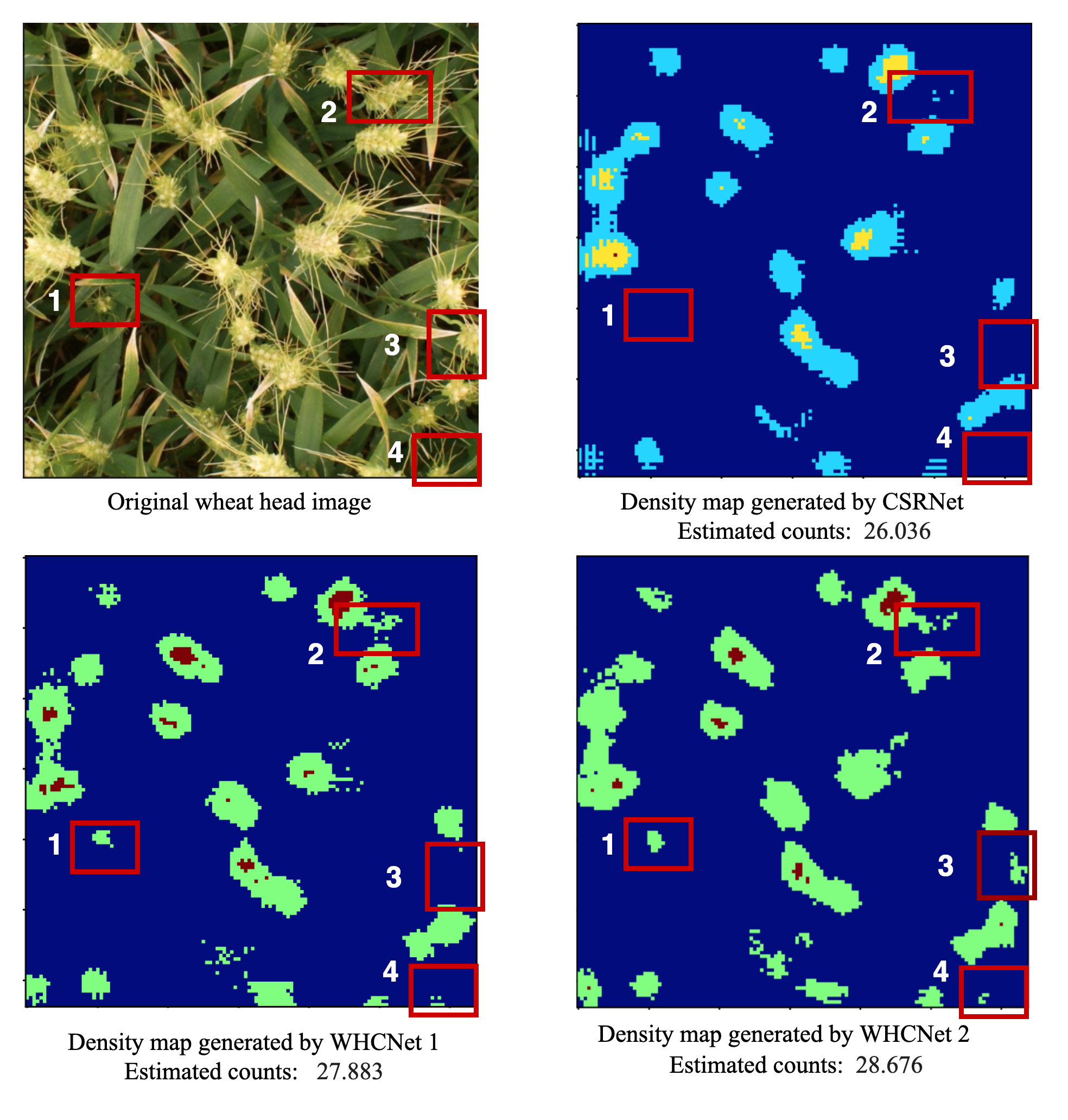}
    \caption{Comparison of density maps generated by CSRNet, WHCNet\_1 and WHCNet\_2. The original wheat head image is one of the 10 test images provided by Kaggle version GWHD. Red squares labels main differences among density maps. Patch 1, Patch 3 and Patch 4 are not detected by the CSRNet model, but detected by our proposed models. The pattern of Patch 2 is more detailed in our proposed models than CSRNet. Patch 3 is detected by WHCNet 2 only.}
    \label{fig:w3}
\end{figure}

\begin{figure*}[hbt!]
    \centering
    \includegraphics[width=0.9\columnwidth]{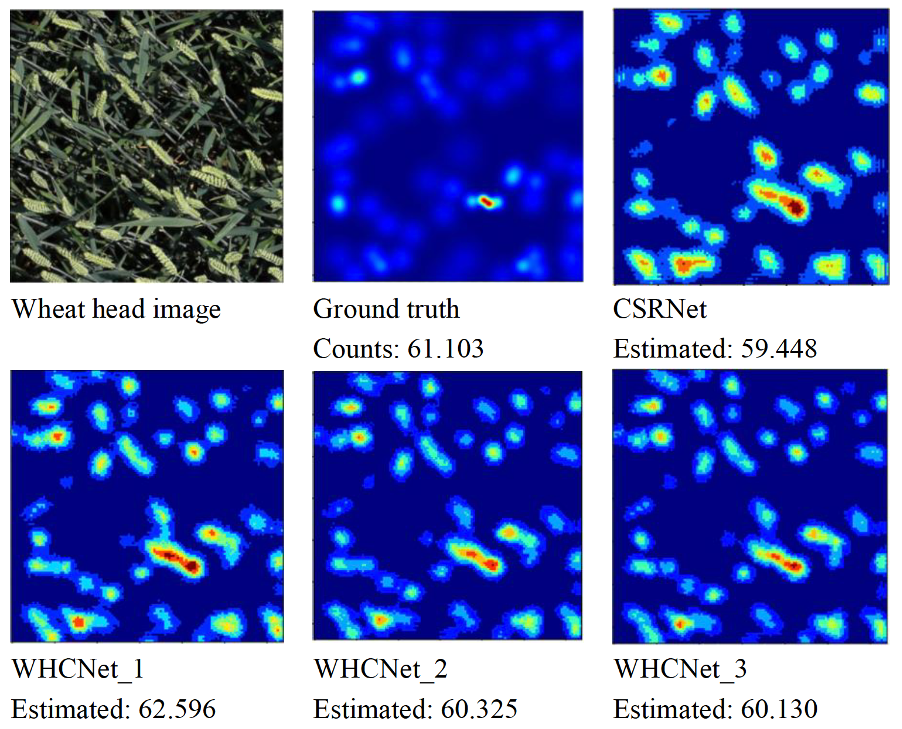}
    \caption{The first row shows one wheat head image and its ground truth density map. The second row presents the generated density map by WHCNet\_1, WHCNet\_2,  WHCNet\_3 and CSRNet. The density maps generated by our proposed models present more detailed information than the baseline model, CSRNet. }
    \label{fig:w4}
\end{figure*}

The comparison of the density maps constructed by our proposed models and CSRNet is illustrated in Figure \ref{fig:w3}. From the figure we can see, our proposed models can detect more detailed information than CSRNet, such as, the small spike in the region of Patch 1 was neglected by CSRNet, but was clearly labeled by our proposed models. The pattern of Patch 2 is more detailed in our proposed models than CSRNet. Moreover, Patch 3 is detected by WHCNet 2 only, because the span of its skip connection is longer than the skip connections in WHCNet\_2, the loss of location information of wheat heads is smaller. While, the architecture of CSRNet is a consecutive CNN, therefore, as the network goes deeper the location information of wheat heads degraded and this affects the quality of the output of density map. Since, there is no ground truth in the test folder of the Kaggle version GWHD, the corresponding ground truth density map is not shown in Figure \ref{fig:w3}. Figure \ref{fig:w4} illustrates another sample of our experiments which includes the ground truth density map in it. 
In conclusion, the quality of density map generated by WHCNet\_2 is the best compared with the original image. The reason is the skip connection in WHCNet\_2 pass the location information to the deeper layers directly, so it is straight forward unlike the braided skip connections in WHCNet\_1.  

\section{Discussion}
Our WHCNets introduced skip connections to the back end to avoid the degradation of the location information, which are represented in lower layers, during the training process of a deep network to improve the quality of the generated density maps. Compared the architecture of the simply stacked layers, skip connections have improved the quality of the density maps through passing the low level features to deeper layers. WHCNet\_1 has achieved the best evaluation scores in terms of MAE and RMSE, but its model size is the biggest in the four models, meanwhile, it needs longer training time as well. The architecture of WHCNet\_1 is complicated including braided skip connections and an down-up sampling part, therefore, to find an optimized model, ablation study is needed. We will explore ablation study in our future work.  

WHCNet\_2 has achieved the best quality of density maps observed from Figure \ref{fig:w3} and Figure \ref{fig:w4}, but its MAE and RSME score were not the best ones. We assume that the reason is complicated and several factors should be taken into consideration: Firstly, our proposed method is a supervised method, thus, the annotation information plays an very important role in the performance of a supervised task. Wheat heads in field plant are often occlude or covered partly by leaves, meanwhile, the environment is very complicated as well, so, its hard to annotated accurately. Secondly, finding a proper ground truth generation algorithm is crucial. In this study, we use the geometry-adaptive kernel, which is designed for dense crowd counting task, however, in our wheat head images, some wheat heads are densely distributed in the image, but some wheat heads are sparsely distributed. Besides, different from crowds, the wheat heads shape is featured with the elongated shape, thus, the geometry-adaptive kernel may not have the ability to delineate the long stripped object. Therefore, a suitable ground truth density map generation method is needed for the wheat head counting task. GWHD has a revised version which took the orientation of wheat heads into account, as the course project time is limited, we will use the revised GWHD version to develop an optimized  ground truth generation algorithm in our further research. Thirdly, data augmentation methods used in this study are simply the cropping and flipping, so the training dataset is limited, hence the model may not generalize well to other situations. We will apply more image augmentation methods to enhance the performance of our models.
Last but not least, since, MAE and RSME are evaluated based on the ground truth density maps and the predicted density maps, the error of the ground truth density maps may cause the error of MAE and RMSE. Therefore, more evaluation methods should be considered in this study.

The performance of WHCNet\_3 dropped slightly compared with  WHCNet\_1 and WHCNet\_2, because its architecture is simple, we could add more layers or increase the number of filters to improve it. Nevertheless, WHCNet\_2 and WHCNet\_3 outperformed the baseline model in terms of MAE, RMSE and model size. 

\section{Conclusion}
In this study, we have proposed three novel models called WHCNet\_1, WHCNet\_2, and WHCNet\_3 for the wheat head counting task and the high-quality density map generation from wheat head images with CNNs. 
We have trained and tested our approaches on GWHD dataset, a large, diverse, and well-labelled dataset of wheat images and built by a joint international collaborative effort. Our WHCNets are composed
of two major components: a CNN as the front-end for wheat head image feature extraction and a skip connected CNN for the back-end to generate high-quality density maps to accomplish the wheat head counting task. We compared our methods with CSRNet, a deep learning
method that can understand highly congested scenes and
perform accurate count estimation as well as present high
quality density maps. By taking the advantage of the skip connections between CNN layers, WHCNets combined low level features, the local information, with high level features to make density map predictions, thus, the density maps can have both high spatial resolution and detailed representations of the input images.
Experiments demonstrated that our methods outperformed CSRNet by the evaluation metrics, MAE and RMSE. 

\bibliographystyle{unsrtnat}
\bibliography{wheat}  






\end{document}